%% file: main.tex
\documentclass{bmvc2k}

\title{FaceGCD: Generalized Face Discovery via Dynamic Prefix Generation}

\addauthor{Yunseok Oh}{oys5339@inha.edu}{1, 2}
\addauthor{Dong-Wan Choi$^\dagger$}{dchoi@inha.ac.kr}{1}

\addinstitution{
 Department of CSE, \\ 
 Inha University,\\
 South Korea
}
\addinstitution{
 AutoLabs, Inc.,\\
 South Korea
}

\runninghead{YS. OH, DW. CHOI}{FaceGCD}

\usepackage{booktabs}
\usepackage{wrapfig}
\usepackage{tcolorbox}
\tcbuselibrary{skins}
\usepackage{graphicx}
\usepackage{xcolor}

\usepackage{multirow}
\usepackage{amsmath}
\usepackage{amssymb}
\usepackage{bm}
\usepackage{multirow} 

\newcommand{\eat}[1]{}

\newcommand{\smalltitle}[1]{ \vspace{1mm}{\noindent\textbf{#1}.\hspace{1mm}}}

\newcommand{\firstbest}[1]{\textbf{#1}}

\newcommand{\secondbest}[1]{\underline{#1}}

\DeclareMathAlphabet{\mathmybb}{U}{bbold}{m}{n}

\begin{document}

\maketitle
\input{abstract}
\input{section_1}
\input{section_2}
\input{section_3}
\input{section_4}
\input{section_5}

\newpage
\bibliography{egbib}

\input{appendix}

\end{document}

%% file: abstract.tex
\begin{abstract}
Recognizing and differentiating among both familiar and unfamiliar faces is a critical capability for face recognition systems and a key step toward artificial general intelligence (AGI). Motivated by this ability, this paper introduces \textit{generalized face discovery} (GFD), a novel open-world face recognition task that unifies traditional face identification with \textit{generalized category discovery} (GCD). GFD requires recognizing both labeled and unlabeled known identities (IDs) while simultaneously discovering new, previously unseen IDs. Unlike typical GCD settings, GFD poses unique challenges due to the high cardinality and fine-grained nature of face IDs, rendering existing GCD approaches ineffective. To tackle this problem, we propose \textit{FaceGCD}, a method that dynamically constructs instance-specific feature extractors using lightweight, layer-wise prefixes. These prefixes are generated on the fly by a \textit{HyperNetwork}, which adaptively outputs a set of prefix generators conditioned on each input image. This dynamic design enables FaceGCD to capture subtle identity-specific cues without relying on high-capacity static models. Extensive experiments demonstrate that FaceGCD significantly outperforms existing GCD methods and a strong face recognition baseline, \textit{ArcFace}, achieving state-of-the-art results on the GFD task and advancing toward open-world face recognition.
\end{abstract}


%% file: section_1.tex
\section{Introduction} \label{sec:intro}





Humans and even some animals appear to possess a natural ability to recognize and differentiate among various faces, including those they have never seen before. Studies such as \cite{innate} show that the ability to identify and group unfamiliar faces as distinct individuals is evident early in the development of infants as well as newborn animals. These findings suggest that the ability to distinguish and categorize known and unknown faces into individual groups is a fundamental aspect of natural intelligence, and thus a critical capability for the pursuit of artificial general intelligence (AGI).



Within this motivation, this paper newly introduces the \textit{generalized face discovery} (GFD) problem, inspired by \textit{generalized category discovery} (GCD) \cite{GCD}. GFD aims to recognize both known and unknown face identities (IDs), requiring not only the classification of known face images but also the discovery of new IDs for unknown face images, as illustrated in Figure 
\begin{wrapfigure}{r}{0.5\textwidth}
  \centering
    \includegraphics[height=5.2cm]{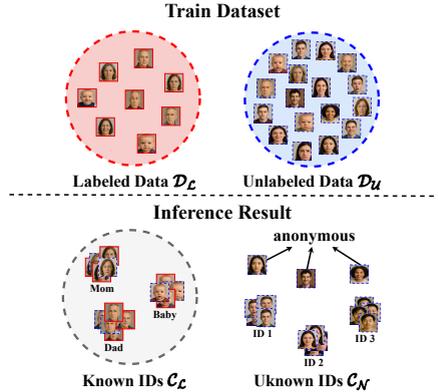}
    \caption{\textbf{Generalized Face Discovery.} Given a training set of labeled and unlabeled face images, GFD aims to classify known IDs and simultaneously discover unknown IDs at inference time.}
    \label{fig:GFD_scenario}
\end{wrapfigure}
\ref{fig:GFD_scenario}. More specifically, given a dataset consisting of labeled face images of known IDs and unlabeled face images corresponding to either known or unknown IDs, our goal is to train a model that can classify known face images into their corresponding IDs while simultaneously discovering new clusters of unknown face images, where each cluster is recognized as a distinct unknown ID.



Despite extensive work on face recognition, existing paradigms each address only a portion of the GFD challenge. \textit{Face identification} ~\cite{arcface, cosface, sphere_face, adacos} is typically trained in a closed-set manner and thus cannot recognize unknown IDs outside the training set. In contrast, \textit{face clustering} ~\cite{lgcn, adanets, lce} is fully unsupervised, grouping only unlabeled face images based on visual similarity without leveraging any prior information about known IDs. Although \textit{open-set face recognition} ~\cite{open-set, open-set-2, open-set-jounal} aims to detect individual novel faces, it treats all unfamiliar faces as outlying instances, lacking the notion of grouping them into distinct IDs. Thus, none of the existing face recognition paradigms fully tackle the mixed scenario of simultaneously recognizing known IDs and discovering new IDs.



\begin{figure}[t]
  \centering
    \includegraphics[width=\textwidth]{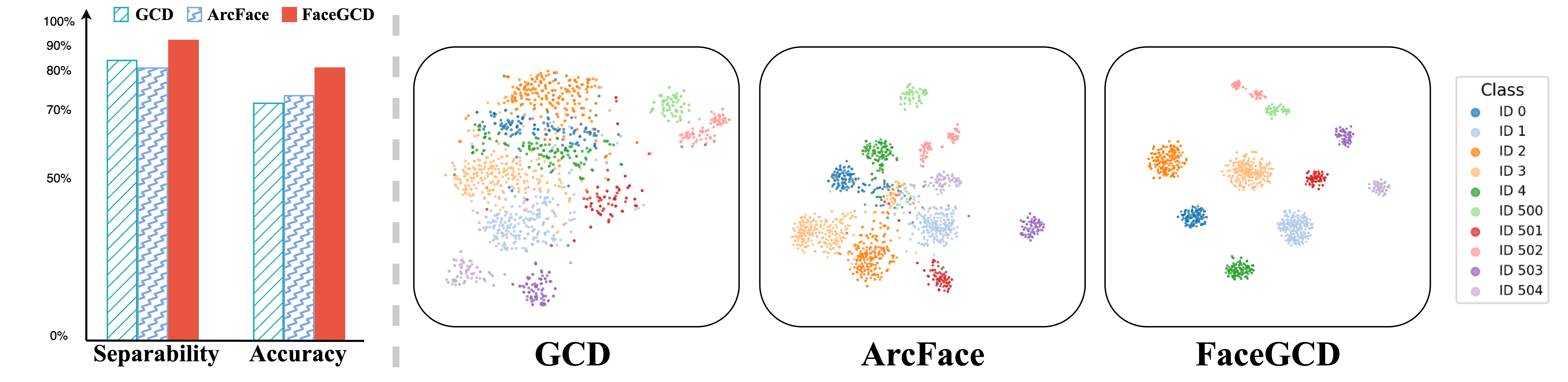}
    \caption{\textbf{(Left)} \textbf{Overall performance.} Separability (defined in Appendix~\ref{appendix:separability_evaluation}) and accuracy on YTF 1000 for GCD \cite{GCD}, ArcFace \cite{arcface}, and FaceGCD (ours). \textbf{(Right) Feature space visualization.} t-SNE features of known (0 - 4) and unknown (500 - 504) IDs.}
    \label{fig:GFD_vis}
\end{figure}

Technically, GFD can be viewed as a specialized form of GCD, where the major difference appears to be only the underlying datasets they focus on. However, this distinction makes it highly challenging to apply existing GCD methods to GFD, as they fail to construct discriminative features on face datasets, as observed in Figure \ref{fig:GFD_vis}. Specifically, GCD methods are typically designed for relatively generic and heterogeneous classes (e.g., CIFAR-100), which are easier to discriminate and far fewer in number. In contrast, GFD involves hundreds or even thousands of visually similar classes (i.e., face IDs), where existing GCD approaches tend to break down. These challenges underscore the need for a tailored solution that can handle the high-cardinality and fine-grained nature of face discovery in an open-world setting.



In this paper, we propose \textit{FaceGCD}, a novel method specifically designed for the GFD problem, capable of capturing highly diverse identity-specific features across numerous fine-grained classes, unlike existing GCD approaches that rely on generic representations. Rather than training a massive feature extractor to this end, FaceGCD dynamically constructs a tailored feature extractor for each input image by augmenting the backbone model with small, layer-wise prefix tokens generated by corresponding prefix generators.  These prefixes~\cite{prefix_tuning} modulate intermediate features at each layer, enabling fine-grained adaptation without significantly increasing model size. To further enhance the adaptability, we employ a \textit{HyperNetwork}~\cite{hypernetwork}-based model that generates prefix generators in an image-specific manner. Given an input, the hypernetwork adaptively produces layer-wise prefix generators, resulting in a feature extractor tailored to capture subtle identity-specific cues. By dynamically adjusting its feature space, FaceGCD achieves finer discrimination among visually similar IDs, thereby overcoming the limitations of static alternatives, as demonstrated in Figure \ref{fig:GFD_vis}.


In our extensive experiments, FaceGCD consistently achieves the best performance on the GFD task, outperforming existing GCD methods by a substantial margin. Moreover, as summarized in Figure \ref{fig:GFD_vis}, it also surpasses a strong face recognition baseline, \textit{ArcFace} \cite{arcface}, confirming its effectiveness in handling both known and unknown IDs. These results indicate that FaceGCD successfully bridges the gap between face identification and clustering, taking a significant step toward open-world face recognition.

%% file: section_2.tex
\section{Related works}
\smalltitle{Face Recognition}
Face recognition has long been a central challenge in computer vision, and \textit{face identification} is one of the most widely studied task. In face identification, a model classifies a given face image into one of a predefined set of known IDs, typically assuming a closed-set scenario where all test IDs are present during training. One of the most popular methods is \textit{ArcFace} \cite{arcface}, which introduces an additive angular margin loss to the traditional softmax classifier, greatly enhancing the discriminative power of the resulting face embeddings. Due to its strong performance and simplicity, ArcFace has become a widely adopted baseline in the areas of face recognition and identification.


Another prominent study of face recognition is \textit{face clustering}, where the goal is to group large collections of unlabeled face images by IDs. With the abundance of unlabeled face images available online, recent approaches have taken advantage of deep learning and graph-based techniques to improve clustering quality. L-GCN~\cite{lgcn} frames clustering as a graph-based link prediction task by constructing local face subgraphs and training a graph convolutional network (GCN) to predict face pairs belonging to the same ID. Similarly, Ada-NETS \cite{adanets} introduces an adaptive neighbor discovery mechanism in a learned embedding space, generating cleaner affinity graphs by selectively pruning noisy edges.


Lastly, \textit{open-set face recognition} relaxes the closed-set assumption by allowing for unknown IDs \cite{open-set}. In this scenario, test images may include faces of subjects not encountered during training. The goal is to accurately recognize known individuals while simultaneously detecting faces that do not match any known ID. Recent approaches \cite{open-set,open-set-jounal,open-set-2} often leverage out-of-distribution detection techniques, such as applying a confidence threshold or similarity-based scoring, to identify unknown faces effectively.


Despite extensive studies in the individual areas above, existing face recognition tasks each address only a partial aspect of the GFD challenge, and none fully handle the joint scenario of simultaneously identifying known IDs and discovering unknown ones. Specifically, face identification excels at recognizing familiar faces but cannot generalize to IDs beyond those known in training. Face clustering can group faces into novel IDs but lacks a mechanism to leverage prior knowledge of known versus unknown IDs. Open-set face recognition can detect unknown IDs but treats them collectively as a single unknown class, without differentiating among new IDs. Thus, existing methods each tackle only a subset of the problem, leaving a critical gap that the proposed GFD scenario aims to address.

\smalltitle{Generalized Category Discovery}
Novel category discovery (NCD) \cite{NCD} addresses the problem of discovering new classes from unlabeled data using limited annotations. In the standard NCD setting, models are trained on labeled data from only known classes and then applied to unlabeled data that exclusively contains unknown classes, assuming a strict separation between known and unknown classes. This constraint is relaxed in the generalized category discovery (GCD) setting \cite{GCD}, which requires simultaneously identifying known classes and discovering novel classes from unlabeled data that may contain both. Existing GCD approaches can be broadly categorized into two-stage methods that apply semi-supervised contrastive learning followed by clustering \cite{GCD, promptcal}, and single-stage methods that jointly train classifiers on known and unknown classes using self-distillation and entropy regularization \cite{simGCD}. Recent advances also explore prompt-based techniques \cite{promptcal} to improve unknown class discovery by enhancing representation adaptability.

However, all previous GCD methods primarily target generic and heterogeneous datasets with a small number of classes, performing not effectively when faced with fine-grained, high-cardinality datasets, to be shown by our experimental results. This limitation underscores the necessity of our proposed method, explicitly tailored to the complexity and scale of real-world face data.


%% file: section_3.tex
\section{Method}

\subsection{GFD Problem}


\smalltitle{Problem Formulation}
Let us formally define our tackled problem, generalized face discovery (GFD). As illustrated in Figure~\ref{fig:GFD_scenario}, the training data consists of both labeled and unlabeled face images. Let $\mathcal{D}_{\mathcal{L}}={(x_i^L, y_i^L)}_{i=1}^{N_L}$ denote the labeled set, where each label $y_i^L$ belongs to a set $\mathcal{C}_{\mathcal{L}}$ of known IDs, and let $\mathcal{D}_{\mathcal{U}}=({x_j^U})_{j=1}^{N_U}$ denote the unlabeled set, where each sample $x_j^U$ may correspond to either to a known ID in $\mathcal{C}_{\mathcal{L}}$ or to an unknown ID in $\mathcal{C}_{\mathcal{N}}$. This mixed version of the unlabeled set generalizes the standard novel class discovery (NCD) setting~\cite{NCD}, where only unknown categories correspond to unlabeled images. Given $\mathcal{D}_{\mathcal{L}}$ and $\mathcal{D}_{\mathcal{U}}$, the goal of GFD is to train a classifier that can assign each test image $x$ to a known ID in $\mathcal{C}_{\mathcal{L}}$ if $x$ belongs to a known ID, or to a newly discovered ID in $\mathcal{C}_{\mathcal{N}}$ if $x$ belongs to a group of similar, other unlabeled faces.

\smalltitle{Technical Challenges}
Unlike generalized category discovery (GCD) \cite{GCD}, which deals with a dataset with a moderate number of heterogeneous categories, GFD presents unique technical challenges due to the extremely fine-grained and high-cardinality nature of face IDs. Face datasets often contain hundreds or even thousands of visually similar IDs, requiring highly discriminative features that can capture subtle variations, beyond the representational capacity of generic feature extractors commonly employed in GCD. Existing GCD model architectures, mostly designed to operates on generic datasets (e.g., ImageNet~\cite{imagenet}), struggle to preserve such fine-grained distinctions in their feature spaces, resulting in poor clustering performance on face data, as illustrated in Figure \ref{fig:GFD_vis}. That said, simply increasing the capacity of the feature extractor is not only inefficient but also ineffective under limited supervised learning setting like GFD. Therefore, it is crucial to devise a more adaptive and efficient mechanism capable of extracting ID-sensitive features without relying on brute-force model scaling.



\subsection{FaceGCD with Dynamic Prefix Generation}
\begin{figure}[t]
   \centering
    \includegraphics[height=6cm]{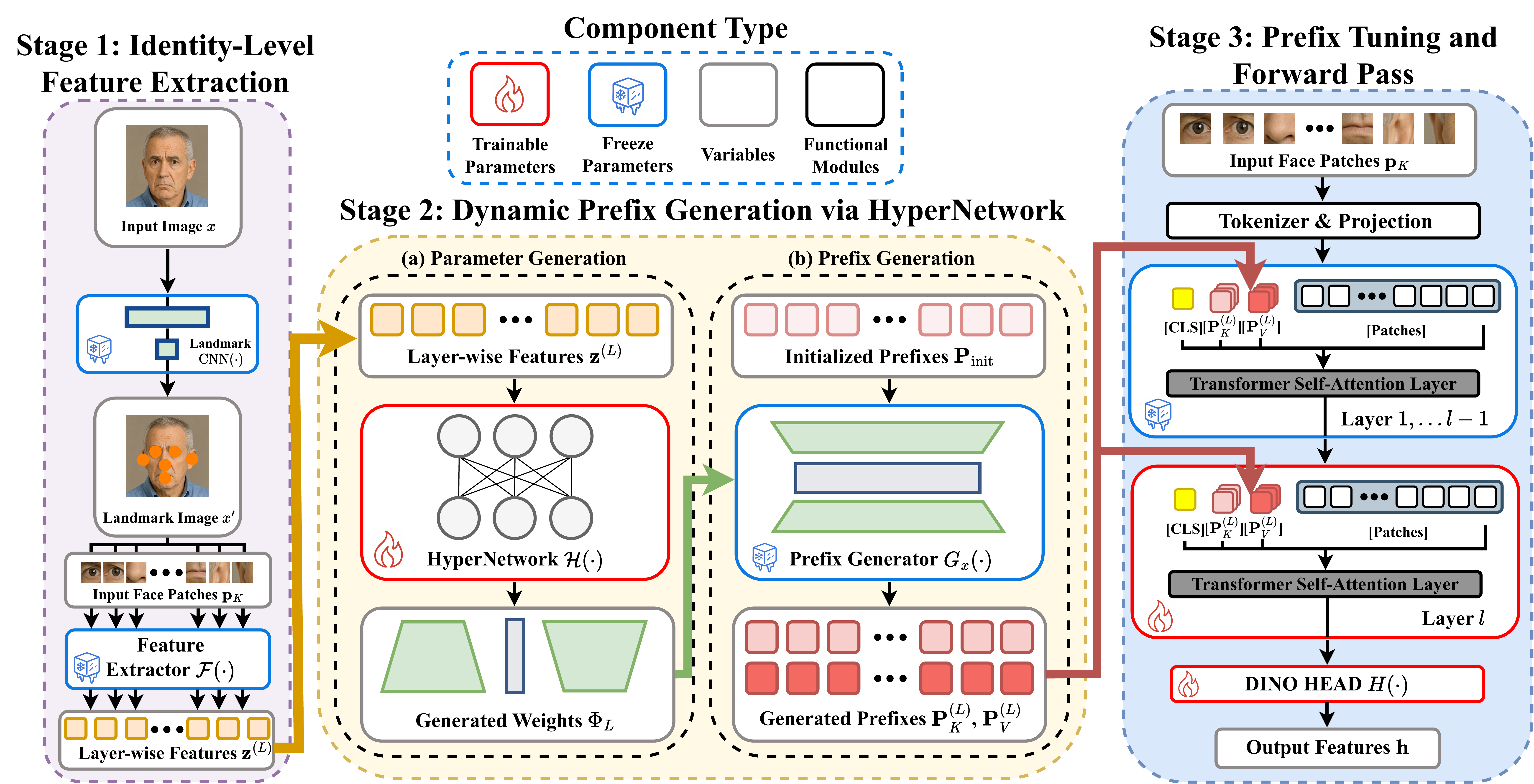}
   \caption{\textbf{FaceGCD architecture overview.}}
   \label{fig:architecture_overview}
\end{figure}



In this paper, we propose FaceGCD, which effectively addresses the aforementioned challenges of GFD by employing our novel architecture that combines prefix tuning~\cite{prefix_tuning, VPT} with a lightweight hypernetwork~\cite{hypernetwork}, as illustrated in Figure \ref{fig:architecture_overview}. 

\smalltitle{Proposed Architecture}
Our core idea is to inject layer-wise prompt tokens~\cite{VPT, prompt_tuning, hyperprompt}, referred to as \textit{prefixes}, into each layer of a pretrained Vision Transformer (ViT)~\cite{vit}. Crucially, these prefixes should not be static but dynamically adapted to each input image, enabling instance-specific modulation of the feature extractor. For this instance-level adaptation, a possible na\"{i}ve approach is to use a prompt pool, which however suffers from the same limitations in existing GCD architectures, to be shown in our ablation study in Table~\ref{table:ablation_study}, unless the pool size is significantly increased. A more effective solution, employed in FaceGCD, is to generate a unique prompt on the fly using a prefix generator, rather than relying on a large prompt pool. That said, our ablation study (Table~\ref{table:ablation_study}) also reveals that prefix generator itself must have a substantial capacity to accommodate the diversity of face images, to the extent that even a 30$\boldsymbol{\times}$ increase in capacity fails to yield competitive performance. Our FaceGCD method addresses this capacity issue by introducing a lightweight hypernetwork, conditioned on the input features, that produces the weights of the prefix generator on the fly. This design eliminates the need for maintaining a large prefix generator, while enabling both instance-specific adaptability and prompt diversity. On top of our proposed architecture, FaceGCD processes the following three stages, each of which is described in detail below.


\smalltitle{Stage 1: Identity-Level Feature Extraction} 
As a common preliminary stage in face recognition, FaceGCD also begins by detecting key landmarks (e.g., eyes, nose, mouth corners) from a given face image $x$ using a pretrained landmark CNN~\cite{part-fvit}, resulting in a landmark-annotated image $x'$. Guided by these landmarks, $x'$ is partitioned into $K$ patches $\{\mathbf{p}_K\}_{K=1}^{k}$, which are then fed into a static feature extractor (a pretrained ViT), yielding layer-wise features $\mathbf{z}^{(L)}$ for $L=\{1,\dots,l\}$ that are shared across images of the same ID. Since this frozen feature extractor has been pretrained on generic data, the resulting identity-level features are expected to retain common and balanced facial knowledge across both known and unknown IDs. These features are therefore used as conditioning inputs for the hypernetwork in the next stage, as indicated by the yellow arrowed line in Figure \ref{fig:architecture_overview}.


\smalltitle{Stage 2: Dynamic Prefix Generation via HyperNetwork}
To achieve instance-specific adaptability without significant computational overhead, we introduce a two-step prefix generation process using a hypernetwork that dynamically produces a prefix generator.

\smalltitle{\textit{(a) Parameter Generation}}
Given the extracted features $\mathbf{z}^{(L)}$ from the previous stage, we employ a hypernetwork $\mathcal{H}(\cdot)$, implemented as a lightweight, trainable 2-layer MLP, which generates layer-specific parameters $\Phi_L = \{\Phi_{(1)},\dots,\Phi_{(l)}\}$. These parameters are then used to build an instance-specific prefix generator $G_x(\cdot)$ for the given image $x$, as depicted by the green arrowed line in Figure \ref{fig:architecture_overview}.


\smalltitle{\textit{(b) Prefix Generation}}
With the resulting prefix generator $G_x(\cdot)$ based on $\Phi^{(L)}$, we create two types of layer-wise prefixes, namely $\mathbf{P}^{(L)}_{K}(x)$ for the keys and $\mathbf{P}^{(L)}_{V}(x)$ for the values in the self-attention module in the backbone ViT model. Based on these two steps (a) and (b), we have: 
$$\mathbf{P}^{(L)}_{K}(x), \mathbf{P}^{(L)}_{V}(x) = G_x(\mathbf{P}_{\text{init}}; \mathcal{H}(\mathbf{z}^{(L)})),$$
where $\mathbf{P}_{\text{init}}$ is a randomly initialized prefix tokens. This two-step approach provides an efficient yet flexible mechanism for generating diverse prefix tokens without requiring to maintain an extensive prompt pool or prefix generator, thereby effectively modulating features at a fine-grained, instance-specific level.


\smalltitle{Stage 3: Prefix Tuning and Forward Pass}
As shown by the red arrowed line in Figure \ref{fig:architecture_overview}, the generated prefixes are dynamically injected into each layer of the backbone ViT. Specifically, for each transformer layer $L$ with its self-attention module having the key, value, and query vectors, denoted by $K^{(L)}$, $V^{(L)}$, and $Q^{(L)}$, respectively, we prepend the prefixes $\mathbf{P}^{(L)}_{K}$ and $\mathbf{P}^{(L)}_{V}$ to the original key and value vectors. In order to partially preserve the original token information~\cite{dualprompt}, we leave the query vectors unchanged. The resulting representations are: 
$$\tilde{K}^{(L)} = [\mathbf{P}^{(L)}_{K}; K^{(L)}], \quad \tilde{V}^{(L)} = [\mathbf{P}^{(L)}_{V}; V^{(L)}], \quad \tilde{Q}^{(L)} = Q^{(L)}, $$
where $\tilde{K}^{(L)}$, $\tilde{V}^{(L)}$, and $\tilde{Q}^{(L)}$ represent the augmented layer-wise attention inputs, enabling the backbone feature extractor to adaptively capture instance-specific features, which are crucial for discriminating among fine-grained and visually similar face IDs. Finally, given this instance-adaptive feature extractor, we apply a semi-supervised $k$-means algorithm (SSK)~\cite{GCD} to the set of representation vectors from the final head $H(\cdot)$, one per image, and assign each image a label, either known or newly discovered, based on the clustering results.


\subsection{Training Objective}

\smalltitle{Trainable Components}
In the proposed architecture shown in Figure \ref{fig:architecture_overview}, we optimize only the hypernetwork $\mathcal{H}(\cdot)$, the final layer of the backbone ViT model, and its DINO head $H(\cdot)$. All other components, including the remaining layers of the backbone, the landmark CNN model, and the static feature extractor, remain frozen during training. Notably, the prefix generators are not directly trained, but instead the hypernetwork is optimized to dynamically generate their parameters. To be shown in our ablation study (see Table \ref{table:ablation_study}), the proposed architecture introduces only a small portion of additional and trainable parameters, resulting in reduced computational overhead and improved training efficiency.

\smalltitle{Loss Function}
To highlight the effectiveness of our architectural design, we adopt the same semi-contrastive loss used in the standard GCD baseline~\cite{GCD}. For each batch $B \subset \mathcal{D}_{\mathcal{U}} \cup \mathcal{D}_{\mathcal{L}}$, our loss function is defined as:
$$
    \mathcal{L}_{B} = (1-\tau) \sum_{i\in B}\mathcal{L}^{u}_{i} + \lambda \sum_{i\in B \cap \mathcal{D}_{\mathcal{L}}}\mathcal{L}^{s}_{i},
$$
where: (1) $\mathcal{L}^{u}_{i}$ is the unsupervised contrastive loss computed for each sample $i$ in the batch, (2) $\mathcal{L}^{s}_{i}$ is the supervised contrastive loss applied only to labeled samples therein, (3) $\tau$ is a temperature scaling hyperparameter, and (4) $\lambda$ is a weighting coefficient balancing the supervised and unsupervised loss terms. The term $\mathcal{L}^{u}_{i}$ encourages consistency between the embedding of each sample and its augmented counterpart, while pushing apart different samples in the batch. The supervised term $\mathcal{L}^{s}_{i}$ additionally brings together samples sharing the same ID label and separates those of different IDs. See Appendix~\ref{appendix:loss} for full formulations.

%% file: section_4.tex
\section{Experiments}
\smalltitle{Datasets}
We evaluate our method on six datasets constructed from two widely-used face recognition benchmarks: YouTube Faces (YTF) ~\cite{youtubefaces} and CASIA-WebFaces (CASIA)~\cite{casia_faces}, each with subsets of 500, 1000, and 2000 IDs. For each dataset, we randomly sample a total of IDs from the full ID set and divide them evenly into known and unknown IDs. Known IDs are partially labeled during training, while unknown IDs remain unlabeled. This setting mirrors real-world face recognition scenarios involving many visually similar IDs. Detailed dataset construction is provided in the Appendix~\ref{appendix:dataset_appendix}.


\smalltitle{Evaluation Metrics}
We follow the standard evaluation metric widely used in the GCD scenario~\cite{GCD,simGCD,promptcal,cms}, where clustering accuracy is computed using the Hungarian optimal assignment algorithm~\cite{kuhn1955hungarian} to match predicted clusters with ground-truth labels. Accuracy is reported separately for \textit{Known}, \textit{Novel}, and \textit{All} IDs. The Hungarian assignment is computed once over the full set of IDs (both known and unknown), and clustering accuracies are derived based on this mapping.

\smalltitle{Implementation Details}
For all ViT-based methods, including ours and GCD baselines~\cite{GCD, simGCD, promptcal, cms}, we use ViT-B/16~\cite{vit} as the common backbone, while ArcFace-based methods~\cite{arcface} use ResNet-101~\cite{resnet}. Both backbones are pretrained on the MS1MV3~\cite{ms1mv3} face dataset: ViT-B/16 with the self-supervised DINO framework~\cite{dino}, and ResNet-101 with supervised ArcFace loss. As a result, ArcFace benefits from label supervision during pretraining, whereas ViT-based models are unaware of ID labels in this phase. For ViT-based models, we follow~\cite{part-fvit} and set the patch size to 8, extracting [CLS] tokens for downstream tasks. All methods utilize a shared facial landmark detector based on MobileNetV3~\cite{mobilenetv3}, as in~\cite{part-fvit}, to extract landmark-guided facial patches. For the static feature extractor in Stage 1 of FaceGCD, we use the same ViT-B/16 architecture as the main backbone but keep it frozen during training. Full implementation details are provided in Appendix~\ref{appendix:implementation_details}.



\subsection{Experimental Results}
\smalltitle{Comparison with GCD Baselines}
We first compare FaceGCD with recent GCD methods~\cite{GCD, simGCD, promptcal, cms} on GFD benchmarks. Table~\ref{table:face_gcd_result} shows the accuracy for All, Known, and Novel IDs across six GFD benchmark datasets constructed from YouTube Faces and CASIA-WebFaces. FaceGCD consistently outperforms all methods across every dataset. Notably, on YouTube Faces 1000, which includes 500 known and 500 unknown IDs, FaceGCD achieves the highest overall accuracy (82.7\%), overwhelmingly surpassing CMS~\cite{cms}, the previous best in the GCD literature. This gain is largely attributed to a substantial improvement in Novel ID accuracy, where FaceGCD outperforms SimGCD~\cite{simGCD} and PromptCAL~\cite{promptcal} by margins of over 8-12 percentages in most cases, while also maintaining competitive accuracy on Known IDs. Performance trends are similar across larger datasets, such as YTF 2000 and CASIA 2000, where the number of IDs and intra-ID variation present greater challenges. Even under these conditions, FaceGCD shows robust scalability, achieving the highest overall accuracy, indicating its effectiveness in handling large-scale, fine-grained identity discovery tasks. These results explain the strength of our instance-specific prefix generation strategy, balancing adaptability to novel IDs and stability for recognition of known ones. Additional results in GCD benchmarks are reported in Appendix~\ref{appendix:gcd_result}.

\begin{table}[t]
\begin{center}
\normalsize
\renewcommand{\arraystretch}{1.15}
\resizebox{\textwidth}{!}{
\begin{tabular}{l|c|cc|c|cc|c|cc|c|cc|c|cc|c|cc}
\toprule
\multirow{2.5}{*}{\centering \textbf{Method}} & \multicolumn{3}{|c}{\textbf{YTF 500}} & \multicolumn{3}{|c}{\textbf{YTF 1000}} & \multicolumn{3}{|c}{\textbf{YTF 2000}} & \multicolumn{3}{|c}{\textbf{CASIA 500}} & \multicolumn{3}{|c}{\textbf{CASIA 1000}} & \multicolumn{3}{|c}{\textbf{CASIA 2000}} \\

\cmidrule(r){2-4} \cmidrule(r){5-7} \cmidrule(r){8-10} \cmidrule(r){11-13} \cmidrule(r){14-16} \cmidrule(r){17-19}
& \textbf{All} & \textbf{Known} & \textbf{Novel} & \textbf{All} & \textbf{Known} & \textbf{Novel} & \textbf{All} & \textbf{Known} & \textbf{Novel} & \textbf{All} & \textbf{Known} & \textbf{Novel} & \textbf{All} & \textbf{Known} & \textbf{Novel} & \textbf{All} & \textbf{Known} & \textbf{Novel} \\

\midrule
GCD~\cite{GCD} & 70.5 & 86.4 & 54.5 & 70.4 & 78.1 & 62.7 & 70.3 & 79.9 & 60.7 & 40.8 & 41.0 & 40.6 & 45.4 & 44.2 & 46.5 & 47.7 & 47.2 & 48.3 \\
SimGCD~\cite{simGCD} & 74.1 & \textbf{96.4} & 51.8 & 72.8 & \textbf{95.1} & 50.1 & 72.6 & \textbf{92.7} & 52.5 & 31.2 & 32.3 & 30.1 & 39.2 & 37.7 & 40.7 & 41.8 & 39.4 & 44.2 \\
PromptCAL~\cite{promptcal} & 75.2 & 88.6 & 61.8 & 75.0 & 86.1 & 63.8 & 66.6 & 76.9 & 56.2 & 44.9 & 43.8 & 45.9 & 49.9 & 49.3 & 50.4 & 40.3 & 39.6 & 41.0 \\
CMS~\cite{cms} & 52.2 & 73.4 & 30.9 & 44.5 & 61.6 & 27.4 & 26.6 & 32.9 & 20.2 & 25.3 & 26.5 & 24.1 & 24.4 & 25.0 & 23.7 & 22.7 & 23.1 & 22.3 \\
\textbf{FaceGCD (Ours)}  & \textbf{81.2} & 93.8 & \textbf{68.6} & \textbf{82.7} & 93.2 & \textbf{72.1} & \textbf{83.6} & 91.8 & \textbf{75.4} & \textbf{58.8} & \textbf{68.9} & \textbf{48.7} & \textbf{52.2} & \textbf{52.3} & \textbf{52.2} & \textbf{56.1} & \textbf{60.4} & \textbf{51.7} \\
\bottomrule
\end{tabular}
}
\end{center}
\caption{\textbf{FaceGCD vs existing GCD methods on various GFD benchmarks.}}
\label{table:face_gcd_result}
\end{table}

\begin{table}[t]
\begin{center}
\normalsize
\renewcommand{\arraystretch}{1.15}
\resizebox{\textwidth}{!}{
\begin{tabular}{l|c|cc | c|cc |c|cc |c|cc |c|cc |c|cc}
\toprule
\multirow{2.5}{*}{\centering \textbf{Feature Extractor}} 
& \multicolumn{3}{c}{\textbf{K-Means}~\cite{k_means}} 
& \multicolumn{3}{|c}{\textbf{DBSCAN}~\cite{dbscan}} 
& \multicolumn{3}{|c}{\textbf{HAC}~\cite{hac}} 
& \multicolumn{3}{|c}{\textbf{L-GCN}~\cite{lgcn}} 
& \multicolumn{3}{|c}{\textbf{Ada-NETS}~\cite{adanets}} 
& \multicolumn{3}{|c}{\textbf{SSK}~\cite{GCD}} \\
\cmidrule(r){2-4} \cmidrule(r){5-7} \cmidrule(r){8-10}
\cmidrule(r){11-13} \cmidrule(r){14-16} \cmidrule(r){17-19}
& \textbf{All} & \textbf{Known} & \textbf{Novel}
& \textbf{All} & \textbf{Known} & \textbf{Novel}
& \textbf{All} & \textbf{Known} & \textbf{Novel}
& \textbf{All} & \textbf{Known} & \textbf{Novel}
& \textbf{All} & \textbf{Known} & \textbf{Novel}
& \textbf{All} & \textbf{Known} & \textbf{Novel} \\
\midrule
ArcFace~\cite{arcface} & 66.5 & 65.4 & 67.6 & 13.9 & 13.9 & 13.9 & 68.5 & 68.7 & 68.4 & 33.2 & 17.7 & 48.7 & 24.6 & 27.2 & 21.9 & 73.0 & 78.3 & 67.6 \\
ArcFace~\cite{arcface} + GCD~\cite{GCD} & 69.6 & 83.9 & 55.3 & 27.5 & 41.0 & 14.0 & 70.1 & 82.8 & 57.3 & 44.6 & 47.2 & 41.9 & 24.9 & 27.7 & 22.0 & 72.0 & 87.9 & 56.1 \\
\textbf{FaceGCD (Ours)} & \textbf{78.6} & \textbf{86.6} & \textbf{70.5} & \textbf{42.0} & \textbf{49.1} & \textbf{34.8} & \textbf{82.0} & \textbf{88.1} & \textbf{75.9} & \textbf{52.3} & \textbf{47.5} & \textbf{57.1} & \textbf{32.6} & \textbf{35.9} & \textbf{29.4} & \textbf{82.7} & \textbf{93.2} & \textbf{72.1} \\
\bottomrule
\end{tabular}
}
\end{center}
\caption{\textbf{FaceGCD vs ArcFace on YTF 1000 with various clustering schemes.}}
\label{table:face_clustering_result}
\end{table}

\smalltitle{Comparison with ArcFace on Clustering Methods}
We additionally examine whether the standard face recognition baseline, ArcFace~\cite{arcface}, performs well in the proposed GFD setting. Since ArcFace is not originally designed for GFD, we consider two variants, namely ArcFace and ArcFace + GCD, where ArcFace indicates the pretrained ArcFace itself and ArcFace + GCD is the fine-tuned version with YTF 1000 using the semi-contrastive loss (as in a typical approach in GCD). Furthermore, in this experiments, we focus on the effectiveness of various clustering algorithms, including standard unsupervised algorithms (K-Means~\cite{k_means}, DBSCAN~\cite{dbscan}, and HAC~\cite{hac}), advanced face clustering techniques (L-GCN~\cite{lgcn} and Ada-NETS~\cite{adanets}), and the SSK method commonly used in GCD~\cite{GCD, promptcal}.

Table~\ref{table:face_clustering_result} summarizes the performance of each model, namely the feature extractor of either FaceGCD or ArcFace, combined with various clustering methods. Across all settings, FaceGCD takes the best performance under every clustering strategy. Although ArcFace-based methods perform reasonably well with K-Means and HAC on known IDs, they struggle to generalize to novel IDs, resulting in decreased overall accuracy. While additional GCD fine-tuning offers slight improvements, the resulting performance remains far below that of FaceGCD, due to its limited ability to discover novel identities.

\subsection{Ablation Studies}
\begin{table}[t]
\begin{center}
\footnotesize
\renewcommand{\arraystretch}{1.2}
\resizebox{\textwidth}{!}{
\begin{tabular}{l|ccc|c|c|c}
\toprule
\textbf{Method} & \textbf{All} & \textbf{Known} & \textbf{Novel} & \textbf{Additional Params} & \textbf{Trainable Params} & \textbf{Total Params} \\
\midrule
\textbf{FaceGCD (Prefix size: 10)} & \textbf{82.9} & 92.9 & \textbf{72.8} & \multirow{3}{*}{13.8M (6.6\%)}  & \multirow{3}{*}{41.1M (19.8\%)}  & \multirow{3}{*}{207.7M}  \\
~~~~Prefix size: 5         & 80.9 (-2.1)         & 91.1 (-1.8) & 70.4 (-2.4)          &   &   & \\
~~~~Prefix size: 20        & 82.7 (-0.2)        & \textbf{93.2 (+0.3)} & 72.1 (-0.7)    &  &  & \\
\midrule
Static Prefix Generator       & 76.9 (-6.0)         & 81.7 (-11.2) & 72.1 (-0.7)         & 399.9M (67.3\%) & 427.2M (72.0\%)  & 593.8M  \\
Static Prefix Pool     & 42.6 (-40.3)        & 52.8 (-40.1) & 32.3 (-40.5)        & \textbf{3.5M (1.8\%)}   & \textbf{30.8M (15.6\%)}  & 197.3M \\
\bottomrule
\end{tabular}
}
\end{center}
\caption{
\textbf{Ablation study on prefix size and alternative prefixing strategies on YTF 1000.}
}
\label{table:ablation_study}
\end{table}
We conduct a detailed ablation study on the YTF1000 benchmark to examine the effectiveness of key components in FaceGCD. Specifically, we analyze the impacts of (i) prefix size, (ii) hypernetwork strategy, and (iii) dynamic prefix generation, respectively. The results are summarized in Table~\ref{table:ablation_study}, and implementation details for each variant are described in Appendix~\ref{appendix:ablation_details}.

\smalltitle{Impact of Prefix Size}
We first investigate the effect of prefix size. Reducing the prefix length to 5 leads to a noticeable drop in overall accuracy (–2.1\%), while increasing it to 20 yields only marginal gains (+0.3\% for known classes, –0.7\% for novel classes). This suggests that, while the model is somewhat sensitive to prefix sizes, performance saturates around a moderate length.


\smalltitle{Effectiveness of HyperNetwork}
We further analyze the effectiveness of the hypernetwork strategy of FaceGCD by comparing to an alternative approach with a large-scale static prefix generator, instead of hypernetwork. In order to cover diverse yet visually similar features in face images, the static prefix generator essentially requires a large capacity (over 10$\boldsymbol{\times}$ more than FaceGCD), but it still underperforms, particularly on known IDs. These results demonstrate that our hypernetwork not only enables instance-specific prefix generation in a highly parameter-efficient manner, but also plays a key role in preserving information about known IDs.


\smalltitle{Static Prefix Pool vs. Dynamic Prefix Generation}
Lastly, we compare our dynamic prefix generation scheme against a static prefix pool, where a fixed set of prefix tokens is trained on the training data and a selected subset is injected during inference. As expected, this strictly static approach fails to capture instance-level subtle variations due to its lack of flexibility, resulting in a sharp drop in accuracy (- 40.3\% overall), with substantial degradation for both known and novel IDs. These large performance gaps highlight the importance of dynamic, instance-specific prefix generation for effectively discriminating among visually-similar face IDs in the GFD setting.

%% file: section_5.tex
\section{Conclusion}
In this paper, we introduced Generalized Face Discovery (GFD), a novel open-world face recognition task that requires identifying both known and unknown identities (IDs). To address the unique challenges of GFD, namely high cardinality and fine-grained visual similarity among face IDs, we proposed FaceGCD, a lightweight yet effective architecture that leverages a hypernetwork to generate dynamic, instance-specific prefixes. Unlike conventional GCD approaches or strong face recognition baselines like ArcFace, FaceGCD achieves superior performance across all GFD benchmarks by adapting the feature extractor on a per-instance basis. Our ablation studies further validate the significance of dynamic prefix generation, demonstrating substantial gains over static alternatives. By enabling fine-grained adaptability without full model retraining, FaceGCD offers a scalable and practical solution for open-world face recognition.

\smalltitle{Acknowledgements}
This work was supported in part by Institute of Information \& Communications Technology Planning \& Evaluation (IITP) grants funded by the Korea government (MSIT) (No.2022-0-00448, Deep Total Recall: Continual Learning for Human-Like Recall of Artificial Neural Networks, No.RS-2022-00155915, Artificial Intelligence Convergence Innovation Human Resources Development (Inha University)), and in part by INHA UNIVERSITY Research Grant. Special thanks to the members of the AutoLabs inc. team for their invaluable support and unwavering encouragement throughout this research.

%% file: appendix.tex
\newpage
\appendix

\setcounter{table}{0}
\setcounter{figure}{0}

\renewcommand{\thetable}{A\arabic{table}}
\renewcommand{\thefigure}{A\arabic{figure}}
\renewcommand{\thesection}{\Alph{section}}

\section*{\centering FaceGCD: Generalized Face Discovery \\ via Dynamic Prefix Generation}

\section*{Appendix}
This appendix provides additional implementation and evaluation details to complement the main paper. We begin with a comprehensive description of the implementation settings for FaceGCD and all baselines, including architectural configurations, hyperparameter setups, and clustering protocols. We then detail the construction of the GFD benchmark datasets, covering dataset composition and identity split strategy. Finally, we include further analyses such as feature separability evaluation and cross-domain performance comparisons on generic GCD benchmarks, offering deeper insight into the effectiveness and generality of our proposed method.
\section{Implementation Details}
\label{appendix:implementation_details}
Pretraining is conducted for 50 epochs using the default hyperparameters of the original DINO and ArcFace implementations. After pretraining, all methods fine-tune only the final layer (keeping the rest frozen) for 200 epochs with a batch size of 128, and the best-performing model on validation (All accuracy) is selected for evaluation. Clustering protocols follow each method’s original setup: GCD and PromptCAL apply the SSK~\cite{GCD} clustering algorithm~\cite{GCD, promptcal}, while others adopt their respective strategies.

Hyperparameters for all baseline methods follow their original papers. FaceGCD also inherits the same setup as GCD~\cite{GCD}, including a prefix size of 20 (as in PromptCAL~\cite{promptcal}). All experiments are conducted on a workstation equipped with four NVIDIA RTX A6000 GPUs, an AMD EPYC 7513 processor (2.6 GHz, 32 cores, 64 threads, 128 MB cache), and 512 GB of Samsung DDR4 ECC REG memory (8 × 64 GB at 3200 MHz). ViT models are implemented using the timm PyTorch library~\cite{timm}.

\subsection{Loss Function Details}
\label{appendix:loss}

We provide the full formulations of the semi-supervised contrastive loss components used in our training objective. The total loss for a batch $B \subset \mathcal{D}_{\mathcal{U}} \cup \mathcal{D}_{\mathcal{L}}$ is given by:

\begin{equation*}
    \mathcal{L} = (1-\tau) \sum_{i\in B}\mathcal{L}^{u}_{i} + \lambda \sum_{i\in B \cap \mathcal{D}_{\mathcal{L}}}\mathcal{L}^{s}_{i},
\end{equation*}
where $\tau$ is a temperature scaling hyperparameter, and $\lambda$ balances the contributions of supervised and unsupervised losses.

\smalltitle{Unsupervised Contrastive Loss.}
The unsupervised term encourages consistency between a sample and its augmented variant while contrasting it against other instances in the batch:
\begin{equation*}
    \mathcal{L}^{u}_{i} = -\log \frac{\exp(\mathbf{z}_{i} \cdot \mathbf{z}'_{i}/\tau)}{\sum_{n} \mathmybb{1}_{[n \neq i]}\,\exp(\mathbf{z}_{i} \cdot \mathbf{z}_{n}/\tau)},
\end{equation*}
where $\mathbf{z}_{i}$ and $\mathbf{z}'_{i}$ are the normalized representations of sample $i$ and its augmented counterpart, respectively.

\smalltitle{Supervised Contrastive Loss.}
For labeled samples, we additionally apply a supervised contrastive loss that encourages intra-class compactness and inter-class separability:
\begin{equation*}
    \mathcal{L}^{s}_{i} = - \frac{1}{\mathcal{N}(i)} \sum_{q\in\mathcal{N}(i)}\log \frac{\text{exp}(\mathbf{z}_{i} \cdot \mathbf{z}_{q}/\tau)}{\sum_{n} \mathmybb{1}_{[n \neq i]}\,\exp(\mathbf{z}_{i} \cdot \mathbf{z}_{n}\,/\,\tau)},
\end{equation*}
where $\mathcal{N}(i)$ denotes the set of samples in the batch that share the same ID label as $i$.

\subsection{FaceGCD Architecture and Hyperparameter Setups}
\label{appendix:main_implementation_details}
\smalltitle{HyperNetwork}
Our hypernetwork architecture adopts a standard multi-layer perceptron (MLP) design, following~\cite{hypernetwork_survey} where MLPs have been shown to perform effectively across a wide range of hypernetwork scenarios. To balance performance with computational efficiency, we use a lightweight 2-layer MLP configuration. As illustrated in Table~\ref{table:hypernetwork_architecture}, the hypernetwork takes as input the layer-wise features extracted from the frozen backbone, yielding $L$ layer-wise features per input image, each of batch size $B$.

Each layer-wise feature is first processed by global average pooling ($\text{AvgPool}$), followed by a linear transformation ($\text{Linear}$), a ReLU activation, and a second linear layer. The resulting vector is then reshaped and passed through an additional average pooling layer. Finally, the output is partitioned into separate weight and bias parameters used to generate the key and value prefixes, as well as the parameters for the down-projection and up-projection layers in the prefix generator.

\begin{table}[t]
\begin{center}
\footnotesize
\resizebox{\textwidth}{!}{%
\begin{tabular}{c|l|l}
\toprule
Process & \textbf{Weights and Biases for Down Projection Generation} & \textbf{Weights and Biases for Up Projection Generation} \\
\midrule
\textbf{Input} & \textbf{Layer-wise features}: $(L, B, 197, 768)$ & \textbf{Layer-wise features}: $(L, B, 197, 768)$ \\
\midrule
Step 1 & AvgPool → $(B, 768)$ & AvgPool → $(B, 768)$ \\
Step 2 & Linear: (768 → 1024)& Linear: (768 → 1024) \\
Step 3 & ReLU & ReLU \\
Step 4 & Linear: (1024 → 16 $\times$ 64) & Linear: (1024 → 64 $\times$ 16) \\
Step 5 & Reshape to $(L, B, 16, 64)$ & Reshape to $(L, B, 64, 16)$ \\
\midrule
\multirow{2}{*}{\textbf{Output}} 
& \textbf{Down Projection Weights}: $(L, B, 16, 64)$ & \textbf{Up Projection Weights}: $(L, B, 64, 16)$ \\
& \textbf{Down Projection Biases (AvgPooled)}: $(L, B, 16)$ & \textbf{Up Biases (AvgPooled)}: $(L, B, 64)$ \\
\bottomrule
\end{tabular}%
}
\end{center}
\caption{\textbf{HyperNetwork process for Key/Value Down and Up projections.}}
\label{table:hypernetwork_architecture}
\end{table}

\begin{table}[t]
\begin{center}
\scriptsize
\begin{tabular}{c|l}
\toprule
Process & \textbf{Key/Value Prefixes Generation} \\
\midrule
\textbf{Input} & 
\textbf{Uniform Initialized Key/Value Prefixes}: $(L, B, K, 12, 64)$  \\
\midrule
Step 1 & Down projection with Weights $(L,B, 16, 64)$ and Biases $(L,B, 16)$\\
Step 2 & ReLU activation\\
Step 3 & Up projection with Weights $(L, B, 64, 16)$, and Biases $(L,B, 64)$\\
\midrule
\textbf{Output}
& \textbf{Key/Value prefix}: $(L, B, K, 12, 64)$ \\
\bottomrule
\end{tabular}%

\end{center}
\caption{\textbf{Prefix generation process using Key and Value projections.} Each projection transforms an initialized prompt via down/up projection to produce instance-layer-wise representations.}
\label{table:prefix_generator_architecture}
\end{table}

\smalltitle{Prefix Generator}
To generate layer-wise prefix tokens for each instance, we utilize the weights and biases produced by the hypernetwork to parameterize a lightweight prefix generator. Specifically, for each transformer layer $L$, we construct separate prefix generators for the key and value vectors using a two-stage projection: a down-projection followed by ReLU activation, and then an up-projection. This process is illustrated in Table~\ref{table:prefix_generator_architecture}.

Formally, let $\mathbf{P}_\tau$ denote the random initialized prefix of shape $m \times h \times d$, where $m$ is the prefix length, $h$ is num of head and $d$ is the per-head dimensionality. The generated key and value prefixes for layer $L$ are given by:
\begin{align*}
\mathbf{P}^{(L)}_{K} = \mathcal{H}^{(L)}_{K}(\mathbf{P}_\tau) = \mathbf{U}^{(L)}_{K}(\text{ReLU}(\mathbf{D}^{(L)}_{K}(\mathbf{P}_\tau))), \\
\mathbf{P}^{(L)}_{V} = \mathcal{H}^{(L)}_{V} (\mathbf{P}_\tau) = \mathbf{U}^{(L)}_{V}(\text{ReLU}(\mathbf{D}^{(L)}_{V}(\mathbf{P}_\tau))),
\end{align*}
where $\mathbf{D}^{(L)}_{K/V} \in \mathbb{R}^{d \times b}$ and $\mathbf{U}^{(L)}_{K/V} \in \mathbb{R}^{b \times h \times d}$ are the down- and up-projection matrices for key and value prefixes. Here, $\mathcal{H}$ denotes the hypernetwork, and $b \ll d$ is a bottleneck dimension that reduces computational overhead.

\smalltitle{Hyperparameter setup}
For all experiments, we set the prefix size to 20 per layer. The model is optimized using stochastic gradient descent (SGD) with a momentum of 0.9 and a weight decay of 2e-5. We employ a cosine learning rate schedule with warm-up to stabilize early training. The base learning rate is initialized at 0.1, with a warm-up learning rate of 1e-5 applied over the first 5 epochs. The total number of training epochs is 200, and the learning rate is annealed following cosine decay, with intermediate decay milestones at epochs 90 and 180, each reducing the learning rate by a factor of 0.1.

We use a contrastive temperature $\tau$ of 1.0 for both supervised and unsupervised objectives. The supervised contrastive loss is weighted by a factor of $\lambda = 0.35$ relative to the unsupervised component, following prior GCD setups. These hyperparameters were selected based on validation performance across multiple datasets and were held fixed throughout all reported experiments. We follow the default training setup used in prior GCD methods, including GCD~\cite{GCD} and PromptCAL~\cite{promptcal}, to ensure a fair and consistent comparison.

\subsection{Clustering Methods Setup}
We apply a variety of clustering algorithms to evaluate the feature representations produced by different models. For all clustering methods, the number of clusters is set to match the ground-truth number of classes (i.e., $|\mathcal{C}_{\mathcal{L}}| + |\mathcal{C}_{\mathcal{N}}|$). Below, we detail the configuration used for each method.

\smalltitle{Semi-Supervised K-Means (SSK)~\cite{GCD}}
We use the semi-supervised $k$-means implementation provided by GCD~\cite{GCD}, where the clustering is guided by labeled instances during training. The model is trained on the training set and evaluated on the test set. We set the maximum number of $k$-means iterations to 500, the number of $k$-means initializations to 10, and explore nearest neighbor settings with \texttt{nb-knn}$\in {10, 20, 100, 200}$.

\smalltitle{K-Means~\cite{k_means}}
We use the standard $k$-means implementation from \texttt{scikit-learn} with the following settings: initialization method = \texttt{k-means++}, maximum iterations = 300, convergence tolerance = 1e-4, and clustering algorithm = \texttt{lloyd}.

\smalltitle{DBSCAN~\cite{dbscan}}
We apply DBSCAN using \texttt{scikit-learn}~\cite{scikit-learn}, performing a grid search over $\texttt{eps}$ values \{0.05, 0.1, 0.15, \dots 10.0\} and $\texttt{min-samples}$ values \{3, 5, 8, 10\}, and report the configuration with the highest clustering accuracy.

\smalltitle{Hierarchical Agglomerative Clustering (HAC)~\cite{hac}}
We use Agglomerative Clustering from \texttt{scikit-learn} with various affinity-linkage combinations. Specifically, we evaluate affinities in {\texttt{euclidean}, \texttt{cosine}} and linkage methods in {\texttt{ward}, \texttt{complete}, \texttt{average}, \texttt{single}}, and report the configuration that yields the highest clustering accuracy.

\smalltitle{L-GCN and Ada-NETS~\cite{lgcn,adanets}}
For L-GCN~\cite{lgcn} and Ada-NETS~\cite{adanets}, we follow the original training protocols using the labeled dataset to train the graph-based or adaptive clustering networks. All hyperparameters are set according to the default values provided in their respective papers.

\subsection{Ablation Implementation Details}
\label{appendix:ablation_details}
All models used in the ablation study are trained using the same procedure and hyperparameters as described in the main implementation section~\ref{appendix:main_implementation_details}, with the prefix size fixed at 10.

\smalltitle{Static Prefix Generator}
The `Static Prefix Generator' model eliminates the hypernetwork and generates prefixes directly using prefix generators. For each transformer layer, a 2-layer MLP receives the corresponding layer-wise feature as input and produces key and value prefixes. The architecture and intermediate dimension (1024) match those of the original method. A group of 10 prefixes (key/value pairs) is generated per layer, enabling direct prefix injection without hypernetwork.

\smalltitle{Static Prefix Pool}
The `Static Prefix Pool' adopts a prefix selection mechanism similar to L2P~\cite{l2p}, but applies it to prefixes rather than prompts. A pool consists of 10 previously trained prefix tokens, from which five are selected per instance at inference time using cosine similarity. These selected prefixes are then concatenated with the input token embeddings and inserted into the transformer. This setup enables reuse of a shared prefix pool, removing the need for dynamic generation.

\section{GFD Dataset Construction}
\label{appendix:dataset_appendix}
To evaluate our method under the Generalized Face Discovery (GFD) scenario, we construct six datasets derived from two widely-used face recognition benchmarks: YouTube Faces (YTF)~\cite{youtubefaces} and CASIA-WebFaces (CASIA)~\cite{casia_faces}. For each dataset, we create subsets containing 500, 1000, and 2000 IDs, resulting in a total of six GFD benchmark datasets.

For the YTF benchmark, we first filter out IDs that contain fewer than 100 face images. From the remaining pool, we randomly sample a particular number of IDs (e.g., 1000 for YTF 1000), and split each ID’s data into 90\% for training and 10\% for testing. In the case of CASIA, since the number of images per ID is relatively small, we select only those IDs with at least 50 face images and apply the same train/test split strategy. After obtaining the filtered dataset, we randomly divide the identities into two equal groups:
\begin{itemize}
    \item Known IDs (50\%), which participate in training under a semi-supervised setting (i.e., only a portion of their images are labeled);
    \item Unknown IDs (50\%), which are entirely unlabeled during training.
\end{itemize}
Specifically, for each known ID, only half of its training images are used as labeled data, while the rest are treated as unlabeled. This construction simulates real-world conditions where only partial identity annotations are available, and many IDs remain unknown.
\begin{table}[t]
    \begin{center}        
    \resizebox{0.8\textwidth}{!}{%
    \begin{tabular}{l|c|c|c|c}
        \hline
        \textbf{Dataset} & \textbf{Known IDs} & \textbf{Unknown IDs} & \textbf{Train Data Num} & \textbf{Test Data Num} \\
        \hline
        YTF 500 & 250 & 250 & 48,089 & 11,779 \\
        YTF 1000 & 500 & 500 & 96,002 & 23,523 \\
        YTF 2000 & 1,000 & 1,000 & 190,248 & 46,615 \\
        CASIA 500 & 250 & 250 & 46,991 & 11,999 \\
        CASIA 1000 & 500 & 500 & 89,508 & 22,867 \\
        CASIA 2000 & 1,000 & 1,000 & 184,432 & 47,114 \\
        \hline
    \end{tabular}
    }
    \end{center}
    \caption{\textbf{GFD Benchmark Datasets.}}
    \label{tab:known_unknown_classes}
\end{table}

Table~\ref{tab:known_unknown_classes} summarizes the number of known and unknown IDs, along with the corresponding number of training and test samples for each dataset. This dataset design reflects practical large-scale face recognition scenarios, such as open-set identification in surveillance or personal photo clustering, where a model must handle both labeled known IDs and unlabeled novel ones.

\section{Additional Analysis}
\subsection{Separability Evaluation}
\label{appendix:separability_evaluation}
\smalltitle{Separability Metric} 
To quantitatively assess the quality of learned representations in terms of cluster separability, we adopt Nearest-Neighbor Consistency (NNC) as our separability metric. NNC measures how well the local neighborhood of an embedding preserves identity, offering an intuitive indicator of clustering suitability.

Formally, given an unlabeled dataset with pseudo-labels assigned (e.g., via clustering), we compute, for each sample, the proportion of its $k$-nearest neighbors that share the same label. Specifically, for an embedding $\mathbf{z}_i$ with assigned label $\hat{y}_i$, we retrieve its $k$ nearest neighbors in the embedding space and count how many of them also have label $\hat{y}_i$. The final metric is obtained by averaging this proportion over all samples in the evaluation set:
\begin{align*}
    \text{NNC} = \frac{1}{N} \sum_{i=1}^{N} \frac{1}{k} \sum_{j \in \mathcal{N}_k(i)} \mathmybb{1}[\hat{y}_j = \hat{y}_i],
\end{align*}
where $\mathcal{N}_k(i)$ denotes the set of $k$ nearest neighbors of sample $i$, and $\mathmybb{1}[\cdot]$ is the indicator function.

Higher NNC values indicate that samples of the same ID are more tightly clustered, which in turn reflects better feature separability. This metric is particularly useful in the GFD scenario, where the goal is to embed both known and unknown IDs in a space conducive to accurate clustering.

\smalltitle{Separability Results}
To evaluate the quality of the learned embedding space, we report both clustering accuracy (ACC) and Nearest-Neighbor Consistency (NNC) in Table~\ref{tab:separability_result}.

FaceGCD achieves the highest performance on both metrics, with an ACC of 82.7\% and an NNC of 91.1\%. Compared to GCD, FaceGCD improves ACC by 12.3 points and NNC by 7.5 points, indicating a much more discriminative feature space. Even against ArcFace, a strong face recognition baseline, FaceGCD outperforms both in clustering quality and neighborhood consistency. These results demonstrate that the instance-specific features produced by our dynamic prefixing mechanism lead to better-separated embeddings, which are crucial for accurately grouping both known and unknown face IDs in the GFD setting.
\begin{table}[t]
\begin{center}
\footnotesize
\resizebox{0.33\textwidth}{!}{ 
\begin{tabular}{l|c|c}
\toprule
\textbf{Method} & \textbf{ACC (\%)} & \textbf{NNC (\%)} \\
\midrule
FaceGCD & \textbf{82.7} & \textbf{91.1} \\
GCD~\cite{GCD} & 70.4 & 83.6 \\
ArcFace~\cite{arcface} & 73.0 & 82.7 \\
\bottomrule
\end{tabular}
}
\end{center}
\caption{\textbf{Clustering Accuracy (ACC) and Nearest-Neighbor Consistency (NNC) on YTF 1000.}}
\label{tab:separability_result}
\end{table}

\subsection{Comparison with GCD Methods on Generic GCD Settings}
\label{appendix:gcd_result}
\begin{table}[t]
\begin{center}
\resizebox{\textwidth}{!}{
\begin{tabular}{l|c|cc|c|cc|c|cc|c|cc|c|cc|c|cc}
\toprule
\multirow{2.5}{*}{\centering \textbf{Method}} & \multicolumn{3}{c|}{\textbf{CIFAR100}} & \multicolumn{3}{c|}{\textbf{ImageNet100}} & \multicolumn{3}{c|}{\textbf{CUB}} & \multicolumn{3}{c|}{\textbf{Stanford Cars}} & \multicolumn{3}{c|}{\textbf{FGVC Aircraft}} & \multicolumn{3}{c|}{\textbf{Herbarium 19}} \\
\cmidrule(r){2-4} \cmidrule(r){5-7} \cmidrule(r){8-10} \cmidrule(r){11-13} \cmidrule(r){14-16} \cmidrule(r){17-19}
& \textbf{All} & \textbf{Old} & \textbf{Novel} & \textbf{All} & \textbf{Old} & \textbf{Novel} & \textbf{All} & \textbf{Old} & \textbf{Novel} & \textbf{All} & \textbf{Old} & \textbf{Novel} & \textbf{All} & \textbf{Old} & \textbf{Novel} & \textbf{All} & \textbf{Old} & \textbf{Novel} \\
\midrule
GCD & 73.0 & 76.2 & 66.5 & 74.1 & 89.8 & 66.3 & 51.3 & 56.6 & 48.7 & 39.0 & 57.6 & 29.9 & 45.0 & 41.1 & 46.9 & 35.4 & 51.0 & 27.0 \\
SimGCD & 80.1 & 81.2 & \firstbest{77.8} & 83.0 & 93.1 & 77.9 & 50.3 & 65.6 & 57.7 & 53.8 & \secondbest{71.9} & 45.0 & 54.2 & 59.1 & 51.8 & \firstbest{44.0} & \firstbest{58.0} & \secondbest{36.4} \\
PromptCAL & 81.2 & 84.2 & 75.3 & \secondbest{83.1} & 92.7 & \secondbest{78.3} & 62.9 & 64.4 & \secondbest{62.1} & 50.2 & 70.1 & 40.6 & 52.2 & 52.2 & \firstbest{52.3} & 37.0 & 52.0 & 28.9 \\
CMS & \secondbest{82.3} & \secondbest{85.7} & \secondbest{75.5} & \firstbest{84.7} & \secondbest{95.6} & \firstbest{79.2} & \firstbest{68.2} & \firstbest{76.5} & \firstbest{64.0} & \secondbest{56.9} & \firstbest{76.1} & \secondbest{47.6} & \firstbest{56.0} & \secondbest{63.4} & \firstbest{52.3} & 36.4 & \secondbest{54.9} & 26.4 \\
\textbf{}{Ours} & \firstbest{83.4} & \firstbest{88.8} & 72.5 & 78.4 & \firstbest{95.7} & 69.6 & \secondbest{64.5} & \secondbest{73.1} & 60.1 & \firstbest{62.5} & 52.1 & \firstbest{67.6} & \secondbest{55.4} & \firstbest{63.5} & 51.3 & \secondbest{42.6} & 44.6 & \firstbest{40.8} \\
\bottomrule
\end{tabular}
}
\end{center}
\caption{\textbf{Comparison of generic GCD benchmarks with GCD methods.}}
\label{table:gcd_result}
\end{table}
\smalltitle{Implementation Details}
Our model used in the GCD experiments are trained following the same procedure and hyperparameter settings as outlined in the ~\ref{appendix:main_implementation_details}. For this setup, we use models pretrained on ImageNet~\cite{imagenet} to ensure consistency with existing GCD benchmarks, and do not use the landmark CNN. All experiments are conducted on commonly used datasets in the GCD literature, including CIFAR-100~\cite{cifar}, ImageNet-100~\cite{imagenet}, CUB-200~\cite{cub}, Aircraft~\cite{aircraft}, Stanford Cars~\cite{stadford_Cars} and Herbarium19~\cite{herbarium} following standard class splits and evaluation protocols~\cite{GCD,simGCD,promptcal,cms}.

\smalltitle{Evaluation Results}
To evaluate the generality of our method beyond the face domain, we apply FaceGCD to standard GCD benchmarks that span diverse and fine-grained image domains. Each dataset is split into Old (known) and Novel (unknown) classes following standard GCD protocols. Table~\ref{table:gcd_result} presents the performance comparison across six datasets, using clustering accuracy (ACC) for All, Old, and Novel classes. In the table, \textbf{bold} and \underline{underline} indicate the best and second-best results, respectively.

Despite being designed specifically for face recognition, FaceGCD demonstrates strong generalization capabilities. It achieves competitive or superior accuracy on all datasets, including state-of-the-art performance on fine-grained datasets like CUB and Stanford Cars. On generic datasets such as CIFAR100 and ImageNet100, FaceGCD performs comparably to or slightly below the best-performing methods (within ~1–2\%), while still maintaining consistent performance across both Old and Novel classes.

These results show that the proposed instance-specific prefix mechanism not only excels in the GFD setting but also generalizes effectively to diverse GCD environments. The ability to generate per-instance modulation without relying on any domain-specific prior (e.g., facial landmarks) highlights the versatility and generality of our approach.

%% file: main.bbl
\begin{thebibliography}{44}
\providecommand{\natexlab}[1]{#1}
\providecommand{\url}[1]{\texttt{#1}}
\expandafter\ifx\csname urlstyle\endcsname\relax
  \providecommand{\doi}[1]{doi: #1}\else
  \providecommand{\doi}{doi: \begingroup \urlstyle{rm}\Url}\fi

\bibitem[Caron et~al.(2021)Caron, Touvron, Misra, J{\'{e}}gou, Mairal, Bojanowski, and Joulin]{dino}
Mathilde Caron, Hugo Touvron, Ishan Misra, Herv{\'{e}} J{\'{e}}gou, Julien Mairal, Piotr Bojanowski, and Armand Joulin.
\newblock Emerging properties in self-supervised vision transformers.
\newblock In \emph{2021 {IEEE/CVF} International Conference on Computer Vision, {ICCV} 2021, Montreal, QC, Canada, October 10-17, 2021}, pages 9630--9640, 2021.

\bibitem[Chauhan et~al.(2024)Chauhan, Zhou, Lu, Molaei, and Clifton]{hypernetwork_survey}
Vinod~Kumar Chauhan, Jiandong Zhou, Ping Lu, Soheila Molaei, and David~A. Clifton.
\newblock A brief review of hypernetworks in deep learning.
\newblock \emph{Artif. Intell. Rev.}, 57, 2024.

\bibitem[Choi et~al.(2024)Choi, Kang, and Cho]{cms}
Sua Choi, Dahyun Kang, and Minsu Cho.
\newblock Contrastive mean-shift learning for generalized category discovery.
\newblock In \emph{{IEEE/CVF} Conference on Computer Vision and Pattern Recognition, {CVPR} 2024, Seattle, WA, USA, June 16-22, 2024}, pages 23094--23104, 2024.

\bibitem[Deng et~al.(2009)Deng, Dong, Socher, Li, Li, and Fei{-}Fei]{imagenet}
Jia Deng, Wei Dong, Richard Socher, Li{-}Jia Li, Kai Li, and Li~Fei{-}Fei.
\newblock Imagenet: {A} large-scale hierarchical image database.
\newblock In \emph{2009 {IEEE} Computer Society Conference on Computer Vision and Pattern Recognition {(CVPR} 2009), 20-25 June 2009, Miami, Florida, {USA}}, pages 248--255, 2009.

\bibitem[Deng et~al.(2018)Deng, Guo, and Zafeiriou]{arcface}
Jiankang Deng, Jia Guo, and Stefanos Zafeiriou.
\newblock Arcface: Additive angular margin loss for deep face recognition.
\newblock \emph{CoRR}, abs/1801.07698, 2018.

\bibitem[Deng et~al.(2019)Deng, Guo, Zhang, Deng, Lu, and Shi]{ms1mv3}
Jiankang Deng, Jia Guo, Debing Zhang, Yafeng Deng, Xiangju Lu, and Song Shi.
\newblock Lightweight face recognition challenge.
\newblock In \emph{2019 {IEEE/CVF} International Conference on Computer Vision Workshops, {ICCV} Workshops 2019, Seoul, Korea (South), October 27-28, 2019}, pages 2638--2646, 2019.

\bibitem[Dosovitskiy et~al.(2021)Dosovitskiy, Beyer, Kolesnikov, Weissenborn, Zhai, Unterthiner, Dehghani, Minderer, Heigold, Gelly, Uszkoreit, and Houlsby]{vit}
Alexey Dosovitskiy, Lucas Beyer, Alexander Kolesnikov, Dirk Weissenborn, Xiaohua Zhai, Thomas Unterthiner, Mostafa Dehghani, Matthias Minderer, Georg Heigold, Sylvain Gelly, Jakob Uszkoreit, and Neil Houlsby.
\newblock An image is worth 16x16 words: Transformers for image recognition at scale.
\newblock In \emph{9th International Conference on Learning Representations, {ICLR} 2021, Virtual Event, Austria, May 3-7, 2021}, 2021.

\bibitem[Ester et~al.(1996)Ester, Kriegel, Sander, and Xu]{dbscan}
Martin Ester, Hans{-}Peter Kriegel, J{\"{o}}rg Sander, and Xiaowei Xu.
\newblock A density-based algorithm for discovering clusters in large spatial databases with noise.
\newblock In \emph{Proceedings of the Second International Conference on Knowledge Discovery and Data Mining (KDD-96), Portland, Oregon, {USA}}, pages 226--231, 1996.

\bibitem[G{\"{u}}nther et~al.(2017)G{\"{u}}nther, Cruz, Rudd, and Boult]{open-set}
Manuel G{\"{u}}nther, Steve Cruz, Ethan~M. Rudd, and Terrance~E. Boult.
\newblock Toward open-set face recognition.
\newblock In \emph{2017 {IEEE} Conference on Computer Vision and Pattern Recognition Workshops, {CVPR} Workshops 2017, Honolulu, HI, USA, July 21-26, 2017}, pages 573--582, 2017.

\bibitem[Ha et~al.(2017)Ha, Dai, and Le]{hypernetwork}
David Ha, Andrew~M. Dai, and Quoc~V. Le.
\newblock Hypernetworks.
\newblock In \emph{5th International Conference on Learning Representations, {ICLR} 2017, Toulon, France, April 24-26, 2017, Conference Track Proceedings}, 2017.

\bibitem[Han et~al.(2019)Han, Vedaldi, and Zisserman]{NCD}
Kai Han, Andrea Vedaldi, and Andrew Zisserman.
\newblock Learning to discover novel visual categories via deep transfer clustering.
\newblock In \emph{2019 {IEEE/CVF} International Conference on Computer Vision, {ICCV} 2019, Seoul, Korea (South), October 27 - November 2, 2019}, pages 8400--8408, 2019.

\bibitem[He et~al.(2016)He, Zhang, Ren, and Sun]{resnet}
Kaiming He, Xiangyu Zhang, Shaoqing Ren, and Jian Sun.
\newblock Deep residual learning for image recognition.
\newblock In \emph{2016 {IEEE} Conference on Computer Vision and Pattern Recognition, {CVPR} 2016, Las Vegas, NV, USA, June 27-30, 2016}, pages 770--778, 2016.

\bibitem[He et~al.(2022)He, Zheng, Tay, Gupta, Du, Aribandi, Zhao, Li, Chen, Metzler, Cheng, and Chi]{hyperprompt}
Yun He, Huaixiu~Steven Zheng, Yi~Tay, Jai~Prakash Gupta, Yu~Du, Vamsi Aribandi, Zhe Zhao, YaGuang Li, Zhao Chen, Donald Metzler, Heng{-}Tze Cheng, and Ed~H. Chi.
\newblock Hyperprompt: Prompt-based task-conditioning of transformers.
\newblock In Kamalika Chaudhuri, Stefanie Jegelka, Le~Song, Csaba Szepesv{\'{a}}ri, Gang Niu, and Sivan Sabato, editors, \emph{International Conference on Machine Learning, {ICML} 2022, 17-23 July 2022, Baltimore, Maryland, {USA}}, pages 8678--8690, 2022.

\bibitem[Howard et~al.(2019)Howard, Pang, Adam, Le, Sandler, Chen, Wang, Chen, Tan, Chu, Vasudevan, and Zhu]{mobilenetv3}
Andrew Howard, Ruoming Pang, Hartwig Adam, Quoc~V. Le, Mark Sandler, Bo~Chen, Weijun Wang, Liang{-}Chieh Chen, Mingxing Tan, Grace Chu, Vijay Vasudevan, and Yukun Zhu.
\newblock Searching for mobilenetv3.
\newblock In \emph{2019 {IEEE/CVF} International Conference on Computer Vision, {ICCV} 2019, Seoul, Korea (South), October 27 - November 2, 2019}, pages 1314--1324, 2019.

\bibitem[Jia et~al.(2022)Jia, Tang, Chen, Cardie, Belongie, Hariharan, and Lim]{VPT}
Menglin Jia, Luming Tang, Bor{-}Chun Chen, Claire Cardie, Serge~J. Belongie, Bharath Hariharan, and Ser{-}Nam Lim.
\newblock Visual prompt tuning.
\newblock In Shai Avidan, Gabriel~J. Brostow, Moustapha Ciss{\'{e}}, Giovanni~Maria Farinella, and Tal Hassner, editors, \emph{Computer Vision - {ECCV} 2022 - 17th European Conference, Tel Aviv, Israel, October 23-27, 2022, Proceedings, Part {XXXIII}}, pages 709--727, 2022.

\bibitem[Kobylkov et~al.(2024)Kobylkov, Rosa-Salva, Zanon, and Vallortigara]{innate}
Dmitry Kobylkov, Orsola Rosa-Salva, Mirko Zanon, and Giorgio Vallortigara.
\newblock Innate face-selectivity in the brain of young domestic chicks.
\newblock \emph{Proceedings of the National Academy of Sciences}, 121\penalty0 (40):\penalty0 e2410404121, 2024.

\bibitem[Krause et~al.(2013)Krause, Stark, Deng, and Fei{-}Fei]{stadford_Cars}
Jonathan Krause, Michael Stark, Jia Deng, and Li~Fei{-}Fei.
\newblock 3d object representations for fine-grained categorization.
\newblock In \emph{2013 {IEEE} International Conference on Computer Vision Workshops, {ICCV} Workshops 2013, Sydney, Australia, December 1-8, 2013}, pages 554--561, 2013.

\bibitem[Krizhevsky et~al.(2009)Krizhevsky, Hinton, et~al.]{cifar}
Alex Krizhevsky, Geoffrey Hinton, et~al.
\newblock Learning multiple layers of features from tiny images.
\newblock 2009.

\bibitem[Kuhn(1955)]{kuhn1955hungarian}
Harold~W Kuhn.
\newblock The hungarian method for the assignment problem.
\newblock \emph{Naval research logistics quarterly}, 2\penalty0 (1-2):\penalty0 83--97, 1955.

\bibitem[Lester et~al.(2021)Lester, Al{-}Rfou, and Constant]{prompt_tuning}
Brian Lester, Rami Al{-}Rfou, and Noah Constant.
\newblock The power of scale for parameter-efficient prompt tuning.
\newblock In Marie{-}Francine Moens, Xuanjing Huang, Lucia Specia, and Scott~Wen{-}tau Yih, editors, \emph{Proceedings of the 2021 Conference on Empirical Methods in Natural Language Processing, {EMNLP} 2021, Virtual Event / Punta Cana, Dominican Republic, 7-11 November, 2021}, pages 3045--3059, 2021.

\bibitem[Li and Liang(2021)]{prefix_tuning}
Xiang~Lisa Li and Percy Liang.
\newblock Prefix-tuning: Optimizing continuous prompts for generation.
\newblock In Chengqing Zong, Fei Xia, Wenjie Li, and Roberto Navigli, editors, \emph{Proceedings of the 59th Annual Meeting of the Association for Computational Linguistics and the 11th International Joint Conference on Natural Language Processing, {ACL/IJCNLP} 2021, (Volume 1: Long Papers), Virtual Event, August 1-6, 2021}, pages 4582--4597, 2021.

\bibitem[Liu et~al.(2017)Liu, Wen, Yu, Li, Raj, and Song]{sphere_face}
Weiyang Liu, Yandong Wen, Zhiding Yu, Ming Li, Bhiksha Raj, and Le~Song.
\newblock Sphereface: Deep hypersphere embedding for face recognition.
\newblock In \emph{2017 {IEEE} Conference on Computer Vision and Pattern Recognition, {CVPR} 2017, Honolulu, HI, USA, July 21-26, 2017}, pages 6738--6746, 2017.

\bibitem[Lloyd(1982)]{k_means}
Stuart~P. Lloyd.
\newblock Least squares quantization in {PCM}.
\newblock \emph{{IEEE} Trans. Inf. Theory}, 28:\penalty0 129--136, 1982.

\bibitem[Maji et~al.(2013)Maji, Rahtu, Kannala, Blaschko, and Vedaldi]{aircraft}
Subhransu Maji, Esa Rahtu, Juho Kannala, Matthew~B. Blaschko, and Andrea Vedaldi.
\newblock Fine-grained visual classification of aircraft.
\newblock \emph{CoRR}, abs/1306.5151, 2013.

\bibitem[Pedregosa et~al.(2011)Pedregosa, Varoquaux, Gramfort, Michel, Thirion, Grisel, Blondel, Prettenhofer, Weiss, Dubourg, Vanderplas, Passos, Cournapeau, Brucher, Perrot, and Duchesnay]{scikit-learn}
F.~Pedregosa, G.~Varoquaux, A.~Gramfort, V.~Michel, B.~Thirion, O.~Grisel, M.~Blondel, P.~Prettenhofer, R.~Weiss, V.~Dubourg, J.~Vanderplas, A.~Passos, D.~Cournapeau, M.~Brucher, M.~Perrot, and E.~Duchesnay.
\newblock Scikit-learn: Machine learning in {P}ython.
\newblock \emph{Journal of Machine Learning Research}, 12, 2011.

\bibitem[Shin et~al.(2023)Shin, Lee, Kim, Baek, Kim, and Koh]{lce}
Junho Shin, Hyo{-}Jun Lee, Hyunseop Kim, Jong{-}Hyeon Baek, Daehyun Kim, and Yeong~Jun Koh.
\newblock Local connectivity-based density estimation for face clustering.
\newblock In \emph{{IEEE/CVF} Conference on Computer Vision and Pattern Recognition, {CVPR} 2023, Vancouver, BC, Canada, June 17-24, 2023}, pages 13621--13629, 2023.

\bibitem[Sibson(1973)]{hac}
Robin Sibson.
\newblock {SLINK:} an optimally efficient algorithm for the single-link cluster method.
\newblock \emph{Comput. J.}, 16:\penalty0 30--34, 1973.

\bibitem[Sun and Tzimiropoulos(2022)]{part-fvit}
Zhonglin Sun and Georgios Tzimiropoulos.
\newblock Part-based face recognition with vision transformers.
\newblock In \emph{33rd British Machine Vision Conference 2022, {BMVC} 2022, London, UK, November 21-24, 2022}, page 611, 2022.

\bibitem[Tan et~al.(2019)Tan, Liu, Ambrose, Tulig, and Belongie]{herbarium}
Kiat~Chuan Tan, Yulong Liu, Barbara Ambrose, Melissa Tulig, and Serge~J. Belongie.
\newblock The herbarium challenge 2019 dataset.
\newblock \emph{CoRR}, abs/1906.05372, 2019.

\bibitem[Vareto et~al.(2023)Vareto, G{\"{u}}nther, and Schwartz]{open-set-2}
Rafael~Henrique Vareto, Manuel G{\"{u}}nther, and William~Robson Schwartz.
\newblock Open-set face recognition with neural ensemble, maximal entropy loss and feature augmentation.
\newblock In \emph{36th {SIBGRAPI} Conference on Graphics, Patterns and Images, {SIBGRAPI} 2003, Rio Grande, RS, Brazil, November 6-9, 2023}, pages 55--60, 2023.

\bibitem[Vareto et~al.(2024)Vareto, Linghu, Boult, Schwartz, and G{\"{u}}nther]{open-set-jounal}
Rafael~Henrique Vareto, Yu~Linghu, Terrance~E. Boult, William~Robson Schwartz, and Manuel G{\"{u}}nther.
\newblock Open-set face recognition with maximal entropy and objectosphere loss.
\newblock \emph{Image Vis. Comput.}, 141:\penalty0 104862, 2024.

\bibitem[Vaze et~al.(2022)Vaze, Han, Vedaldi, and Zisserman]{GCD}
Sagar Vaze, Kai Han, Andrea Vedaldi, and Andrew Zisserman.
\newblock Generalized category discovery.
\newblock In \emph{{IEEE/CVF} Conference on Computer Vision and Pattern Recognition, {CVPR} 2022, New Orleans, LA, USA, June 18-24, 2022}, pages 7482--7491, 2022.

\bibitem[Wah et~al.(2011)Wah, Branson, Welinder, Perona, and Belongie]{cub}
Catherine Wah, Steve Branson, Peter Welinder, Pietro Perona, and Serge Belongie.
\newblock The caltech-ucsd birds-200-2011 dataset.
\newblock 2011.

\bibitem[Wang et~al.(2018)Wang, Wang, Zhou, Ji, Gong, Zhou, Li, and Liu]{cosface}
Hao Wang, Yitong Wang, Zheng Zhou, Xing Ji, Dihong Gong, Jingchao Zhou, Zhifeng Li, and Wei Liu.
\newblock Cosface: Large margin cosine loss for deep face recognition.
\newblock In \emph{2018 {IEEE} Conference on Computer Vision and Pattern Recognition, {CVPR} 2018, Salt Lake City, UT, USA, June 18-22, 2018}, pages 5265--5274, 2018.

\bibitem[Wang et~al.(2022{\natexlab{a}})Wang, Zhang, Zhang, Wang, Lin, Zhang, and Sun]{adanets}
Yaohua Wang, Yaobin Zhang, Fangyi Zhang, Senzhang Wang, Ming Lin, YuQi Zhang, and Xiuyu Sun.
\newblock Ada-nets: Face clustering via adaptive neighbour discovery in the structure space.
\newblock In \emph{The Tenth International Conference on Learning Representations, {ICLR} 2022, Virtual Event, April 25-29, 2022}, 2022{\natexlab{a}}.

\bibitem[Wang et~al.(2019)Wang, Zheng, Li, and Wang]{lgcn}
Zhongdao Wang, Liang Zheng, Yali Li, and Shengjin Wang.
\newblock Linkage based face clustering via graph convolution network.
\newblock In \emph{{IEEE} Conference on Computer Vision and Pattern Recognition, {CVPR} 2019, Long Beach, CA, USA, June 16-20, 2019}, pages 1117--1125. Computer Vision Foundation / {IEEE}, 2019.

\bibitem[Wang et~al.(2022{\natexlab{b}})Wang, Zhang, Ebrahimi, Sun, Zhang, Lee, Ren, Su, Perot, Dy, and Pfister]{dualprompt}
Zifeng Wang, Zizhao Zhang, Sayna Ebrahimi, Ruoxi Sun, Han Zhang, Chen{-}Yu Lee, Xiaoqi Ren, Guolong Su, Vincent Perot, Jennifer~G. Dy, and Tomas Pfister.
\newblock Dualprompt: Complementary prompting for rehearsal-free continual learning.
\newblock In Shai Avidan, Gabriel~J. Brostow, Moustapha Ciss{\'{e}}, Giovanni~Maria Farinella, and Tal Hassner, editors, \emph{Computer Vision - {ECCV} 2022 - 17th European Conference, Tel Aviv, Israel, October 23-27, 2022, Proceedings, Part {XXVI}}, pages 631--648, 2022{\natexlab{b}}.

\bibitem[Wang et~al.(2022{\natexlab{c}})Wang, Zhang, Lee, Zhang, Sun, Ren, Su, Perot, Dy, and Pfister]{l2p}
Zifeng Wang, Zizhao Zhang, Chen{-}Yu Lee, Han Zhang, Ruoxi Sun, Xiaoqi Ren, Guolong Su, Vincent Perot, Jennifer~G. Dy, and Tomas Pfister.
\newblock Learning to prompt for continual learning.
\newblock In \emph{{IEEE/CVF} Conference on Computer Vision and Pattern Recognition, {CVPR} 2022, New Orleans, LA, USA, June 18-24, 2022}, pages 139--149, 2022{\natexlab{c}}.

\bibitem[Wen et~al.(2023)Wen, Zhao, and Qi]{simGCD}
Xin Wen, Bingchen Zhao, and Xiaojuan Qi.
\newblock Parametric classification for generalized category discovery: {A} baseline study.
\newblock In \emph{{IEEE/CVF} International Conference on Computer Vision, {ICCV} 2023, Paris, France, October 1-6, 2023}, pages 16544--16554, 2023.

\bibitem[Wightman()]{timm}
Ross Wightman.
\newblock Pytorch image models.
\newblock \url{https://github.com/rwightman/pytorch-image-models}.

\bibitem[Wolf et~al.(2011)Wolf, Hassner, and Maoz]{youtubefaces}
Lior Wolf, Tal Hassner, and Itay Maoz.
\newblock Face recognition in unconstrained videos with matched background similarity.
\newblock In \emph{The 24th {IEEE} Conference on Computer Vision and Pattern Recognition, {CVPR} 2011, Colorado Springs, CO, USA, 20-25 June 2011}, pages 529--534, 2011.

\bibitem[Yi et~al.(2014)Yi, Lei, Liao, and Li]{casia_faces}
Dong Yi, Zhen Lei, Shengcai Liao, and Stan~Z. Li.
\newblock Learning face representation from scratch.
\newblock \emph{CoRR}, abs/1411.7923, 2014.

\bibitem[Zhang et~al.(2023)Zhang, Khan, Shen, Naseer, Chen, and Khan]{promptcal}
Sheng Zhang, Salman~H. Khan, Zhiqiang Shen, Muzammal Naseer, Guangyi Chen, and Fahad~Shahbaz Khan.
\newblock Promptcal: Contrastive affinity learning via auxiliary prompts for generalized novel category discovery.
\newblock In \emph{{IEEE/CVF} Conference on Computer Vision and Pattern Recognition, {CVPR} 2023, Vancouver, BC, Canada, June 17-24, 2023}, pages 3479--3488, 2023.

\bibitem[Zhang et~al.(2019)Zhang, Zhao, Qiao, Wang, and Li]{adacos}
Xiao Zhang, Rui Zhao, Yu~Qiao, Xiaogang Wang, and Hongsheng Li.
\newblock Adacos: Adaptively scaling cosine logits for effectively learning deep face representations.
\newblock In \emph{{IEEE} Conference on Computer Vision and Pattern Recognition, {CVPR} 2019, Long Beach, CA, USA, June 16-20, 2019}, pages 10823--10832, 2019.

\end{thebibliography}
